\documentclass[letterpaper, 10 pt, conference]{ieeeconf}
\IEEEoverridecommandlockouts

\usepackage{booktabs}
\usepackage{amsmath}
\usepackage{amssymb} 

\makeindex                                                           
\usepackage{graphicx}
\DeclareGraphicsExtensions{.pdf,.jpeg,.png}

\usepackage{siunitx}
\usepackage{tikz}
\usetikzlibrary{calc}

\usepackage{mathtools}
\DeclareMathOperator*{\argmax}{arg\,max\;}
\DeclareMathOperator{\logm}{logm}

\usepackage{color}

\usepackage{layouts}  
\newcommand{\TVResultNumberKeptSubjects}{\num[round-precision=0]{11}}

\newcommand{\TVResultCompAccEEGCombined}{\num{0.7786}}

\newcommand{\TVResultCompAccEEGStatement}{\num{0.7576}}

\newcommand{\TVResultCompAccEEGObservation}{\num{0.6318}}

\newcommand{\TVResultCompAccEEGObservationStart}{\num{0.5344}}

\newcommand{\TVResultCompAccEEGObservationEnd}{\num{0.5922}}

\newcommand{\TVResultCompAccButton}{\num{0.9249}}

\newcommand{\TVResultRankDistDiffEEGCombinedInst}{\num{0.0732}}

\newcommand{\TVResultRankNdcgOneEEGCombinedInst}{\num{0.8224}}

\newcommand{\TVResultRankNdcgThreeEEGCombinedInst}{\num{0.8407}}

\newcommand{\TVResultRankDistDiffEEGObservationInst}{\num{0.1324}}

\newcommand{\TVResultRankNdcgThreeEEGObservationInst}{\num{0.6772}}

\newcommand{\TVResultRankNdcgThreeEEGStatementInst}{\num{0.8209}}

\newcommand{\TVResultRankNdcgOneButtonInst}{\num{0.8121}}

\newcommand{\TVResultRankNdcgThreeButtonInst}{\num{0.8152}}

\newcommand{\TVResultRankDistDiffButtonFeat}{\num{0.0528}}

\newcommand{\TVResultRankNdcgOneButtonFeat}{\num{0.8686}}

\newcommand{\TVQuestReferenceAgreementVAMean}{\num{0.6874}}

\newcommand{\TVQuestReferenceAgreementBlocksCombinedAgreeFraction}{\SI[round-precision=0]{42}{\percent}}

\newcommand{\TVQuestReferenceAgreementBlocksCombinedStrongAgreeFraction}{\SI[round-precision=0]{35}{\percent}}

\newcommand{\TVQuestReferenceAgreementBlocksCombinedPositiveFraction}{\SI[round-precision=0]{78}{\percent}}

\begin{document}

\title{Learning User Preferences for Trajectories from Brain Signals}
\author{Henrich Kolkhorst, Wolfram Burgard and Michael Tangermann
  \thanks{ All authors are with the Department of Computer Science,
    University of Freiburg, Germany.  H.K., W.B. and M.T. are with the
    Autonomous Intelligent Systems group and H.K. and M.T. are also
    with the Brain State Decoding Lab.\newline
    Corresponding author's email: kolkhorst@informatik.uni-freiburg.de
  } }
\maketitle

\begin{abstract}Robot motions in the presence of humans should not only be
  feasible and safe, but also conform to human preferences.
        This, however, requires user feedback on the robot's behavior.   In this work, we propose a novel approach to leverage the user's
  brain signals as a feedback modality in order to decode the judgment
  of robot trajectories and rank them according to the user's
  preferences.   We show that brain signals measured using electroencephalography
  during observation of a robotic arm's trajectory as well as in
  response to preference statements are informative regarding the
  user's target trajectory. Furthermore, we demonstrate that user
  feedback from brain signals can be used to reliably infer pairwise
  trajectory preferences as well as to retrieve the target
  trajectories of the user with a performance comparable to explicit
  behavioral feedback.
\end{abstract}

\begin{tikzpicture}[remember picture, overlay]
  \node at ($(current page.south) + (0,0.5in)$) {The International
    Symposium on Robotics Research (ISRR), Hanoi, Vietnam, October
    2019};
\end{tikzpicture}

\section{Introduction}
\label{sec:introduction}

In the vicinity of humans, it is not only important what a robot does,
but also \emph{how} these actions are performed. Especially in the
context of robotic assistants, trajectories should not only be
feasible and free of obstacles, but also comply with the user's
preferences. However, preferences over trajectories may vary between
users, environments, tasks and also time, which poses challenges to
design general cost functions. Instead, it can be beneficial to learn
preferences directly from the user.

To learn preferences, input from the user in the form of
demonstrations or feedback on candidate trajectories is
necessary. Particularly for robots with multiple degrees of freedom or
for impaired users, providing trajectory demonstrations may be
prohibitive. Giving feedback on the robot's behavior, however, is
possible and---especially for relative instead of absolute
ratings---does not require expert knowledge.

In order to obtain the human judgment of a trajectory, different
modalities such as screen-based rating or speech are
conceivable. Brain signals as a feedback modality can be desirable
because their measurement does not interfere with the primary task
and---especially in the context of robotic assistants for impaired
users---can potentially be recorded from users who cannot reliably
control robots through other modalities. However, signals measured
using noninvasive electroencephalography (EEG) typically have an
unfavorable signal-to-noise ratio that makes it challenging to utilize
this information based on single brain responses.

\begin{figure}[t]
  \centering
    \includegraphics[width=\columnwidth]{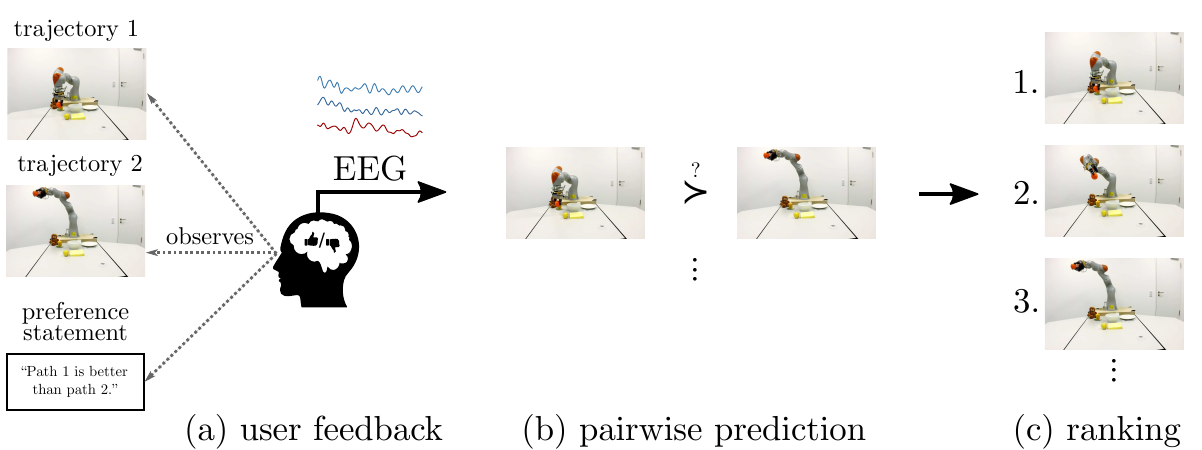}
  \caption{Overview of our approach: (a) We measure the brain response
    of users during trajectory observation and in response to
    preference statements in a pairwise comparison setting using
    electroencephalography (EEG). (b) We predict pairwise preferences
    from the EEG data utilizing methods from Riemannian geometry. (c)
    We combine these predictions to rank the trajectories.}
  \label{fig:schematic}

\end{figure}

In this work, we address the problem of learning trajectory
preferences from EEG-based user feedback. For this, we let the user
observe multiple trajectory pairs and additionally present
a---potentially incorrect---preference statement that indicates which
of the observed trajectories was supposedly better. This offers two
informative sources for discriminative brain signals: As depicted in
Fig.~\ref{fig:schematic}, we classify the user's brain signals during
robot observation and after showing preference statements to predict
the actual pairwise preferences of the user. Subsequently, we rank
trajectories across pairs using the EEG-based predictions. To
facilitate a quantitative evaluation that is comparable across users,
we asked them to give feedback according to common target trajectories
rather than personal preferences in this work. We compare our approach
against explicit user feedback collected from button presses after the
preference statements.

Our main contributions are: 1) We propose an approach to estimate user feedback to trajectories by
classifying brain signals both during trajectory observation and in
reaction to preference statements. 2) We show in experiments with \TVResultNumberKeptSubjects{}
participants, who observed videos of real robot trajectories, that
these brain signals are informative about trajectory judgment and that
our approach enables reliable decoding of the preferred trajectory in
pairwise comparisons in an offline manner. 3) We demonstrate that these pairwise preference predictions based on
brain signals can be used to successfully identify the target
trajectory from the observed ones with a performance that is
comparable to explicit feedback in the form of button presses.

\newpage
\section{Related Work}
\label{sec:related-work}

In order for robots to operate successfully and adequately in human
environments, motion planning should incorporate the user's
demands. In the past, the modeling of human preferences has been
addressed using cost functions
\cite{dragan14integrating,busch17postural} and by integrating
hand-crafted costs into planning \cite{mainprice11planning}. The use
of demonstrations allows to learn reward functions, e.g., from navigation behavior
\cite{kretzschmar16socially} or corrective actions for robot
manipulators \cite{bajcsy13learning}.

Rather than modeling costs manually or providing near-optimal
demonstrations, behavioral feedback can also be used to improve
robotic actions. For example, legibility can be optimized by
minimizing the time until the user can press a button corresponding to
the anticipated goal of the robot \cite{busch17learning}. While
absolute ratings of robotic actions would be desirable for ranking a
set of options, it is easier for human users to give relative feedback
by comparing a small set of items. Relative feedback has been used to
model, e.g., human perception of ``naturalness'' of robot
configurations \cite{jeon18configuration} or preferences in simulated
driving \cite{sadigh17active}. While many approaches assume label
noise in the user feedback \cite{akrour14programming,sadigh17active},
it is typically small and assumed to vary based on the reward
differences. In contrast, strong measurement noise is typically
encountered when using EEG signals. Closely related to our setting is
the work by Jain~et~al.~\cite{jain15learning}, which aims at learning
preferences based on human feedback. However, they utilize
screen-based reranking of trajectories and kinesthetic teaching as
user feedback.

Brain--computer interfaces (BCIs) based on EEG signals are typically
either driven by mental imagery of the user \cite{burget17acting} or
external stimuli \cite{chen15highspeed,schreuder11listen}. In reactive
attention-based BCIs, discriminative information can be obtained from
differing responses---mostly in the form of event-related potentials
(ERPs)---to stimulus classes, such as identifying ``surprising''
outlier stimuli or stimulus changes \cite{schreuder11listen}. Due to
the low signal-to-noise ratio, prediction is typically limited to
binary classification. A large body of different classification
approaches specifically tailored to EEG signals exists
\cite{lotte18review}. State-of-the-art results increasingly utilize
methods from Riemannian geometry
\cite{yger17riemannian,barachant13classification}: Assuming that
relevant information about the mental state in a given time interval
can be represented by the covariance matrix, the corresponding
manifold structure suggests using non-Euclidean distance measures
between data points. Generally, classification performance heavily
depends on the experimental task---implying the mental states to be
classified---and varies from user to user.

Many common BCIs are based solely on the appearance or the identity of
an attended visual stimulus, whereas feedback on trajectories requires
the decoding of the user's \emph{judgment} of an action in a
context. Error-related potentials \cite{chavarriaga14errare}, i.e.,
brain responses to committed or observed errors, form one common type
of brain responses useful for classification. Error-related
potentials are interesting since they are based on judgment of
behavior, but typically also require its fast comprehension by the
user.\\
Brain activity from time windows that are not time-aligned to
specific stimuli can also be informative---e.g., to predict workload
\cite{schultze-kraft16unsupervised} or upcoming task performance
\cite{meinel16pre}. Although desirable for applications, predictive
performance in such asynchronous settings is typically lower than in
stimulus-aligned ones.

Recent work has incorporated brain responses to robotic behavior,
e.g., for error recognition \cite{salazar-gomez17correcting} or
learning gesture mappings with reward signals from EEG
\cite{kim17intrinsic}. In these contexts, the human judgment of the
behavior could be performed near-instantaneously, greatly easing the
temporal alignment of EEG signals for classification. However, in the
case of observing more complex behavior (e.g., trajectories that have
an identical start configuration but continuously deviate thereafter),
the temporal alignment---i.e., the time point the human realizes one
behavior is superior or inferior---cannot easily be inferred.

\section{Decoding Trajectory Preferences from Brain Signals}
\label{sec:approach}

\begin{figure}[t]
  \centering
  \includegraphics[width=\columnwidth]{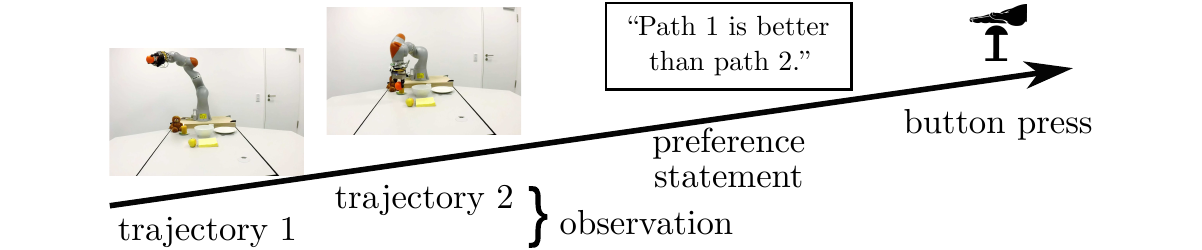}
  \caption{Course of pairwise trajectory comparison: The user observes
    videos of the robot executing two trajectories. Subsequently, we
    present the user with a (potentially incorrect) \emph{preference
      statement} of the form ``Path 1 was better than path 2.'' (or
    vice versa). For evaluation purposes, after a delay we also ask
    the user to press a button if she agrees with the statement.
  }
  \label{fig:pairwise-comp}

\end{figure}

We consider the problem of identifying a user's preferred trajectory
based on feedback by the user: Given a set of trajectories
$\Xi = \lbrace \xi_i \rbrace, i = 1, \ldots, N$, we want to find a
\emph{target trajectory}
$\xi^{*} \in \argmax_{\xi \in \Xi} R_{h}(\xi)$ that maximizes the
(unknown) reward $R_h(\xi)$ of a human user in a given environment. In
order to estimate $\xi^{*}$, we opt to query the user for relative
preference feedback and use this feedback---from EEG or from explicit
button presses for comparison---to rank trajectories.

After obtaining feedback, our approach consists of two prediction
modules, which are depicted in Fig.~\ref{fig:schematic}: First, we
propose to classify pairwise feedback to trajectories from brain
signals during robot observation and in response to preference
statements as described in Section~\ref{sec:decod-pairw-pref}. We
utilize the covariance representations of signals, using methods from
Riemannian geometry. Second, we use this feedback to rank trajectories
both in an instance-based and a feature-based manner (see
Section~\ref{sec:traj-rank-based}). Note that training of both modules
is independent, i.e., classifiers for EEG decoding are not specific to
different environments or preferences and only depend on the user.

In order to obtain pairwise preference feedback from a user, we
perform a sequence of trajectory comparisons
$\mathcal{C} = \lbrace c_1, \ldots, c_{N_c} \rbrace$. Each comparison
$c_j$ consists of two trajectories $\xi_{i_{j,1}}, \xi_{i_{j,2}}$ that
are presented to the user (i.e.,
$c_j= \lbrace i_{j,1}, i_{j,2} \rbrace$).\\
To obtain comparative judgment of the two trajectories, we propose to
use \emph{preference statements}: After presenting a pair of
trajectories, we show a textual statement to the user in the form of
``Path 2 is better than path 1.''  The correctness of the statement
implies a preferred trajectory and should also induce differing
decodable mental states. For control, we subsequently ask the user to
press a button if she judges this statement to be correct (c.f.,
Fig.~\ref{fig:pairwise-comp}).  Hence, for each comparison $c_j$, we
observe a behavioral response by the user in response to the statement
(i.e., a button press if the statement is correct), which implies the
user's pairwise preference.

\subsection{Decoding Pairwise Preferences from Brain Signals}
\label{sec:decod-pairw-pref}

We have two potentially informative sources of user feedback based on
brain responses in each comparison: the brain activity during watching
the trajectory execution by the robot---the \emph{observation}
setting---and the response to a subsequent \emph{statement} (e.g.,
``Path 1 is better than path 2.''). As discussed in
Section~\ref{sec:related-work}, the latter synchronous case is better
suited for classification using brain signals, whereas the former
asynchronous one would be desirable to collect information passively
during fluent human--robot interaction. We train separate classifiers
for both types of signals as well as a combined one. As input, we
extract features from fixed-sized windows of the continuous
frequency-filtered EEG signal.

\subsubsection{Segmentation and Labeling of Brain Signals}
\label{sec:eeg-segmentation}

 For the \emph{observation} setting, we use a \SI{2}{\second}
window temporally centered in the trajectory execution, leading to
feature vectors
$X^{o}_{j,k} \in \mathbb{R}^{N_\mathrm{ch} \times
  N_s^o}$ for trajectory $\xi_{i_{j,k}}$ in comparison
$c_j$ ($k \in \lbrace 1, 2 \rbrace$). Here, $N_\mathrm{ch}$ denotes the
number of channels in the recording and $N_s^o$ the
duration of the window in samples. The temporal centering encodes our
hypothesis that the user judgment evolves, with increasing confidence,
during the observation and---with a high uncertainty of the user at
the start and a low one at the end---intermediate signals might
capture discriminative mental states.

We assume supervision in the form of a training dataset for which the
user's preference---the target trajectory $\xi^*$---is known. While we
aim to infer the pairwise preference in each comparison, such a
comparative judgment is only possible after having observed both
trajectories. Hence, comparative labels are not applicable for
observation windows.  Instead, we opt to use the similarity between
observed candidate trajectories and the target trajectory. As a
trajectory similarity measure, we use a geometric distance: For
trajectory $\xi$, we temporally normalize trajectory durations to
$[0, 1]$ and interpolate trajectory waypoints using cubic splines to
get a trajectory representation $f_\xi(t)$, which we use to calculate
the distance
\begin{equation}
  \label{eq:traj-dist}
  d(\xi_1, \xi_2) = \int_0^1 \left\| f_{\xi_1}(t) - f_{\xi_2}(t)
  \right\|_2 \,\mathrm{d}t .
\end{equation}

Equipped with this distance, we can calculate
$d_\mathrm{target}(\xi_i) = d(\xi_i, \xi^*)$ for each trajectory in
our training data. We obtain binary labels $y^o_{j,k}$ for windows
$X^o_{j,k}$ by thresholding $d_\mathrm{target}(\xi_{i_{j,k}})$ on the
median of all distances in the training data. Other absolute
trajectory metrics or ratings could be used alternatively. Given the
predicted distance class $\hat{y}_{j,k}^{o}$ for each of the two
trajectories in comparison $j$ ($k \in \lbrace 1, 2 \rbrace$), we
derive the pairwise-preferred trajectory $\hat{i}_{j}$ based on the
smaller of the two.

For the \emph{statement} response, we extract a \SI{1}{\second} window
$X^{s}_j \in \mathbb{R}^{N_\mathrm{ch} \times N_s^s}$ for each
comparison $c_j$, starting with the onset of the statement. As a
label, we use the correctness of each statement
$y_j^{s} \in \lbrace\mathit{correct}, \mathit{erroneous}\rbrace$ based
on $d_\mathrm{target}$ of the trajectories. Since the statements are
comparative, the predicted correctness $\hat{y}_j^{s}$ implies the
pairwise-preferred trajectories $\hat{i}_j$. In the rare cases in
which the behavioral response of the user did not correspond to our
label (i.e., the button press implied a preference for the trajectory
further away from the target), we assumed that the user's brain state
also reflected the behavioral judgment and consequently corrected
these labels.

We also \emph{combine} both feedback types in addition to classifying
observation and statement windows separately. For this, we concatenate
the predictions for the statement and the observation windows
$(\hat{y}_j^s \,, \hat{y}^o_{j,k} \,, \hat{y}^o_{j,k'})$ on the comparison
level ($k, k' \in \lbrace 1, 2 \rbrace$), where each prediction is the
output of the separately trained classifiers for each window
type. Note that we order the observation-based predictions such that
the first coincides with the supposedly better trajectory in the
statement. For the combined setting, we use the labels $y_j^s$ from
the corresponding statements to train a logistic regression
classifier.

\subsubsection{Feature Extraction and Classification of Brain Signals}
\label{sec:eeg-features}

We use covariance-based features for the classification of both
observation and statement windows. Since we expect time-locked
event-related potentials only after the statement (c.f.,
Section~\ref{sec:related-work}), we perform baselining and
augmentation for these windows: We baseline each of the statement
windows by subtracting the channelwise average activity in the
\SI{200}{\milli\second} preceding the window and augment them with
prototype responses \cite{barachant14plug}. As prototypes
$P^{+}, P^{-}$, we use the mean response for each class in the
training data. Additionally, we reduce the channel count by projecting
the data using $W^{+}, W^{-} \in \mathbb{R}^{3 \times N_\mathrm{ch}}$
obtained by selecting three components per class based on the largest
eigenvalues from an xDAWN decomposition \cite{rivet09xdawn}. This
leads to augmented statement windows
\begin{equation*}
 \! \tilde{X^{s}_j} \! = \!
  \begin{pmatrix}
    W^{+} P^+,  W^{-} P^-,  W^{+} X_j^s,  W^{-} X_j^s
  \end{pmatrix}^T \!\!\!\in  \mathbb{R}^{12 \times  N_{\mathrm{samples}}}.
\end{equation*}

For both feedback types ($X^{o}_{j,k}$ or $ \tilde{X^{s}_j}$), we
calculate window-wise covariances $C$ using a Ledoit-Wolf
regularization. To account for the symmetric positive-definiteness and
the corresponding undesirable properties of the Euclidean distance
\cite{yger17riemannian}, we project each of the covariance
representations into the corresponding tangent space at the Frech\'et
mean $C_\mathrm{ref}$ of the training samples of the corresponding
window type (c.f., \cite{kolkhorst18guess,barachant13classification}):
\begin{equation}
  \label{eq:tangentspace}
  S = \logm\left( C_\textrm{ref}^{-1/2} C
    C_\textrm{ref}^{-1/2} \right)
\end{equation}
Here, $\logm$ denotes the logarithm of a diagonalizable matrix (i.e.,
the logarithm of each element of the diagonal after the corresponding
decomposition). The upper triangular entries of these projected
covariances $S$ are used as input to the classifier. Due to the
typically small number of training samples that are available per
user, we use $L_2$-regularized logistic regression classifiers.

\subsection{Trajectory Ranking based on Noisy Pairwise Preferences}
\label{sec:traj-rank-based}

In order to rank a set of trajectories based on pairwise preferences,
we follow two conceptual approaches: In the \emph{instance-based}
setting, ranking is performed solely based on the identity of the
compared trajectories (i.e., the index $i$), disregarding any
additional (geometric) information about the
trajectory. Alternatively, learning to rank can be based on (e.g.,
geometric) trajectory features
$\phi: \Xi \to \mathbb{R}^{N_\mathrm{feat}}$.

While only the latter \emph{feature-based} approach allows to rank
trajectories for which no user feedback has been observed, it depends
on the expressiveness of the feature representations $\phi$. Hence, we
propose both an instance-based and a feature-based ranking approach
for combining the pairwise predictions to identify preferred
trajectories.

For \emph{instance-based} ranking solely based on the pairwise
comparison outcome, we use a modified \emph{Borda counting} method
\cite{shah18simple}: For each trajectory $\xi_i$ (which is a candidate
in the comparisons
$J(i) = \lbrace j=1, \ldots, N_c,\; i \in c_j\rbrace$), we count the
number of comparisons in which $\xi_i$ is the predicted pairwise
preference, yielding $\hat{R}_\mathrm{instance}(\xi_i) = \sum_{j \in J(i)}
  \mathbf{1}\lbrace \hat{i}_j = i\rbrace $. Here
$\mathbf{1}\lbrace\cdot\rbrace$ denotes the indicator function that
returns 1 iff the argument is true.

While conceptually simple, the method is asymptotically optimal for
retrieving the most highly ranked items from noisy user observations
\cite{shah18simple}. However, in the case of EEG-based pairwise
preferences, we also have to account for substantial noise in the
predictions, which typically differs between comparisons yet should be
correlated to the classifier's predicted probability $\hat{y}$. We
propose a heuristic extension to the Borda counting that weighs
comparisons based on the confidence
$\mathit{conf}(\hat{y}_j) = | \hat{y}_j - 0.5|$:
\begin{equation}
  \label{eq:reward-borda-count-conf}
  \hat{R}_\mathrm{instance (conf)}(\xi_i) = \sum_{j \in J(i)}
    \mathit{conf}(\hat{y}_j)  \mathbf{1}\lbrace \hat{i}_j
  = i\rbrace 
\end{equation}

For the \emph{feature-based} ranking approach, we follow the common
assumption that the reward is linear in the (geometric) feature
representation $\phi$ of a trajectory
\cite{sadigh17active,jain15learning}:
\begin{equation}
  \label{eq:reward-feature}
  \hat{R}_\mathrm{feat}(\xi_i) = \boldsymbol{\theta}^T \phi(\xi_i)
\end{equation}
Hence, the goal is to find a
$\boldsymbol{\theta} \in \mathbb{R}^{N_\textrm{feat}}$ such that a
pairwise preference of $\xi_m$ over $\xi_n$ in comparison $j$ implies
$\boldsymbol{\theta}^T\phi(\xi_m) >
\boldsymbol{\theta}^T\phi(\xi_n)$. Consequently, we want the projected
feature differences
$\boldsymbol{\theta}^T \left( \phi(\xi_m)-\phi(\xi_n) \right)$ to be
positive and can use them to train a binary classifier
\cite{joachims02optimizing}. As labels, we use the predicted pairwise
preference relations $\mathbf{1} \lbrace \hat{i}_j = m \rbrace$ based on the brain signals of the user (c.f.,
Section~\ref{sec:eeg-segmentation}).

For the feature representation $\phi$ we include geometric information
on the robot movement and on the environment interaction:
For the movement, we include the end effector's mean squared velocity,
mean squared acceleration and mean and maximal squared jerk as well as
mean and maximal joint velocities over the trajectory. For the
environment interactions, we use the features proposed in
\cite{jain15learning} (e.g., minimal distances to scene objects and
distance from the goal), yielding a total of $N_\mathrm{feat}=120$
features. Note that also other---potentially more
discriminative---feature representations could easily be used instead.
We train a $L_2$-regularized logistic regression classifier on this
data and use the classifier predictions to rank trajectories.

\section{Experiments}
\label{sec:experiments}

In order to evaluate our approach for learning trajectory preferences
from brain signals, we assessed the performance of different feedback
types for predicting pairwise preferences as well the inference of
target trajectories based on combining these predictions into a
ranking. We performed the evaluation using data from experiments with
\TVResultNumberKeptSubjects{} participants. Specifically, we wanted to
answer the following questions: (1) Are brain signals as a feedback
modality---passively during trajectory observation or reactively after
a preference statement---informative about user preferences? (2) Is it
possible to classify responses to single comparisons in order to
predict the pairwise-preferred trajectory? (3) Can we use these
predictions to select a trajectory that is close to the user's target
trajectory?

The evaluation design requires a balancing of the diversity of
(preference) trajectories against reproducibility, reduction of
confounders and comparability across participants. To have identical
trajectory execution for all participants, we opted to record videos
of a Kuka iiwa robotic arm executing trajectories and showed these
videos to the participants. While there is a mismatch between watching
videos and observing a robot in the scene, we believe that directly
observing the robot would likely be more immersive, and thus might
lead to even stronger brain responses. Preference statements could
also easily be presented by the robot similar to our setting (e.g.,
using audio). The recording quality of EEG in the proximity of a robot
still allows successful decoding \cite{kolkhorst18guess}.

While it would be desirable to let participants give feedback
according to their personal preference trajectories, this can lead to
infeasible desired target trajectories (especially for novice users)
and differing class distributions in the pairwise preference
prediction task, hindering comparability of performance. Crucially, a
geometric representation is not available for personal preference
trajectories, which precludes a quantitative evaluation of the
predicted preference trajectory based on geometric proximity to the
ground truth. Therefore, we presented \emph{reference} trajectories to
the participants and instructed them to give preference feedback
according to these references. To validate the adequacy of the
references, we also asked participants to rate the personal agreement
with the reference trajectories.

\subsection{Experimental Setup}
\label{sec:data-characteristics}

\begin{figure}[t]
  \centering
  \includegraphics[width=\columnwidth]{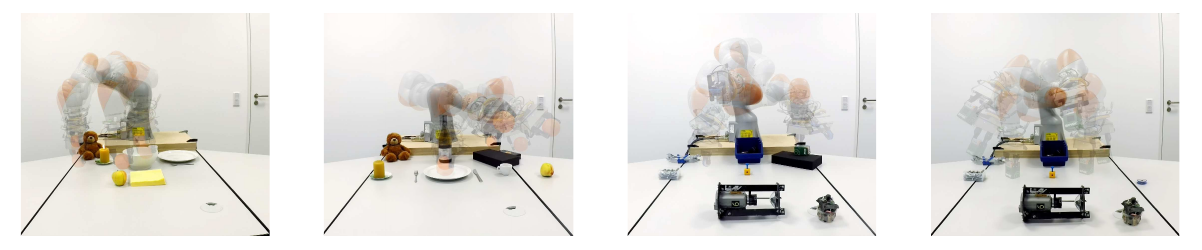}

  \caption{Examples for the different tabletop environments used in
    our experiments. To give an intuition about the robot's path, we
    overlaid screenshots of intermediate configurations.}
  \label{fig:video-examples}

\end{figure}

Each experiment session consisted of \num{16} \emph{preference tasks}
with differing target trajectories. With nine comparisons per target
trajectory, each participant observed a total of \num{288} videos in
\num{144} comparisons.\\
We recorded videos in four different environments with differing
objects (resembling assembly or dinner settings as depicted in
Fig.~\ref{fig:video-examples}). In each environment, there were two
different pairs of initial and final end-effector poses. For each of
these pairs, two dissimilar reference trajectories were selected
(e.g., passing on different sides of an obstacle). Each was the target
of a single preference task. Hence, all trajectories in a task had
identical initial and final end-effector poses and only differed in
the intermediate configurations (please see the supplemental video for
examples). The trajectory videos had a mean duration of
\SI[separate-uncertainty,multi-part-units=single]{5 \pm
  2}{\second}. In order to resemble a realistic robotic setup, a
minority of videos also contained ``failures'' (e.g., the arm touches
scene objects or an item is dropped from the gripper). We sampled
candidate trajectories using RRT-Connect \cite{kuffner00rrtconnect}
and selected a subset to assure variance within tasks.

For each preference task (i.e., an experimental block with a single
target trajectory), we first showed the reference
trajectory. Subsequently, we performed nine comparisons, each
consisting of two videos, followed by a preference statement as
depicted in Fig.~\ref{fig:pairwise-comp}. We only prompted for the
button response \SI{2}{\second} after the appearance of the statement
to reduce a possible influence of motor activity on the brain
response. The order of trajectories in the statement was randomized
and statements were balanced with respect to the expected behavioral
response of the participants. Including pauses between videos and
comparisons, one preference task took approximately 7 minutes.

We acquired the brain signals using a cap holding Ag/AgCl gel-based
passive electrodes positioned according to the extended 10--20 system
with a nose reference. Channel impedances were kept below
\SI{20}{\kilo\ohm}. The amplifier sampled the EEG signals at
\SI{1}{\kilo\hertz}. We used $N_\mathrm{ch}=32$ channels whose signals
were frequency filtered to a band of \SIrange{0.5}{40}{\hertz}. For
the statement responses, we resampled the data to \SI{100}{\hertz}.

We conducted experiment sessions with 14 participants. Following the
declaration of Helsinki, we received approval by the local ethics
committee and obtained written informed consent from
participants. Participants familiarized themselves with the setting in
five comparisons that were not analyzed.

In data from three of the sessions, we observed a large fraction of
artifacts after the statements (more than 30\% of windows $X^s$
exceeded a max--min difference of \SI{100}{\mu\V} in any channel),
likely caused by eye blinks. While the EEG data of these sessions
still contains discriminative information, classification would likely
primarily be based on muscular artifacts rather than brain
signals. Hence, we only kept data from the other
\TVResultNumberKeptSubjects{} participants. We did not reject any data
of the remaining participants, enabling identical sample counts and
ranking tasks. We trained separate classifiers for each participant
and evaluated them in a chronological 5-fold cross-validation.

\subsection{Results for Pairwise Preference Prediction}
\label{sec:pairwise-comparison}

\begin{figure}[t]
  \centering
  \includegraphics[width=\columnwidth]{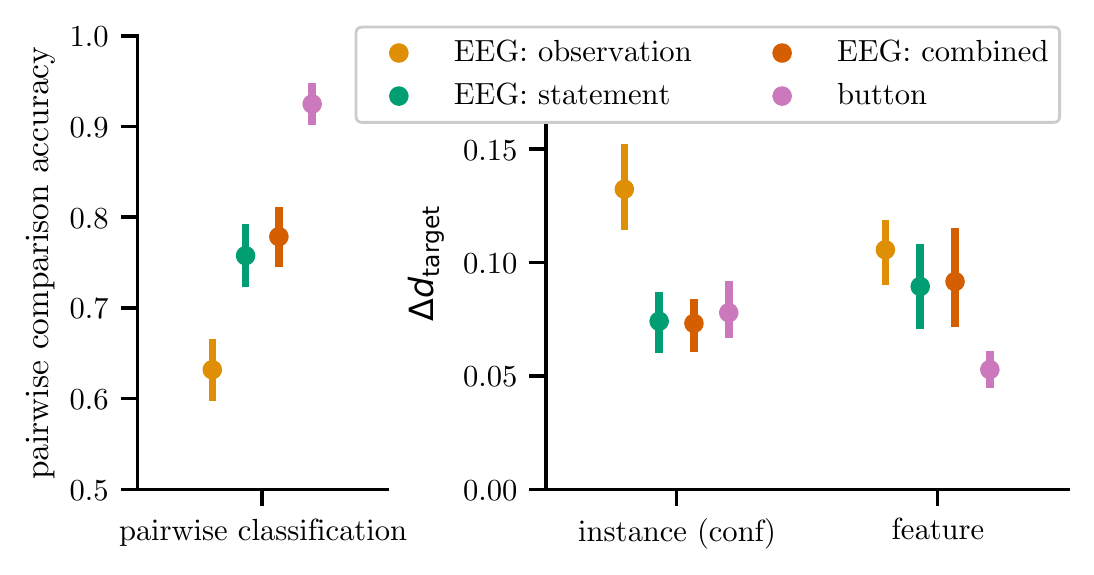}
  \vspace{-3mm}
  \caption{Mean accuracies for all \TVResultNumberKeptSubjects{}
    participants of predictions on a comparison level (left, higher is
    better) and distance differences $\Delta d_\textrm{target}$ from
    the target for the top-ranked trajectory for each preference
    (right, lower is better). Error bars correspond to bootstrapped
    95\% confidence intervals.}
  \label{fig:pred-results}
\vspace{-2mm}
\end{figure}

Before analyzing ranking performance, we discuss the intermediate
pairwise preference prediction.
We evaluated it based on the \emph{comparison accuracy} using the
trajectory with a smaller $d_\mathrm{target}$ as the ground truth,
which assures identical labels for all participants. Hence, also the
behavioral response of the participants did not achieve a perfect
score (which is in line with the assumption of noisily rational
behavior). Nevertheless, the mean accuracy based on button presses
(\TVResultCompAccButton{}, see Fig.~\ref{fig:pred-results}) indicated
that the participants followed the task.

We asked the last nine participants after the experiment to indicate
whether the reference paths agreed with their personal preference. On
a visual analog scale from ``little'' (\num{0}) to ``much'' (\num{1}),
participants marked an average of \TVQuestReferenceAgreementVAMean{},
with seven of nine indicating a tendency to agree (upper half of
scale). When asked after each task, participants stated that reference
trajectory matched their preference in
\TVQuestReferenceAgreementBlocksCombinedPositiveFraction{} of cases
(\TVQuestReferenceAgreementBlocksCombinedAgreeFraction{} ``agree'' and
\TVQuestReferenceAgreementBlocksCombinedStrongAgreeFraction{} ``strong
agree'').

\subsubsection{Electrophysiology of Statement Responses}
\label{sec:res-avg-response}

\begin{figure}[t]
  \centering
  \includegraphics[width=\columnwidth]{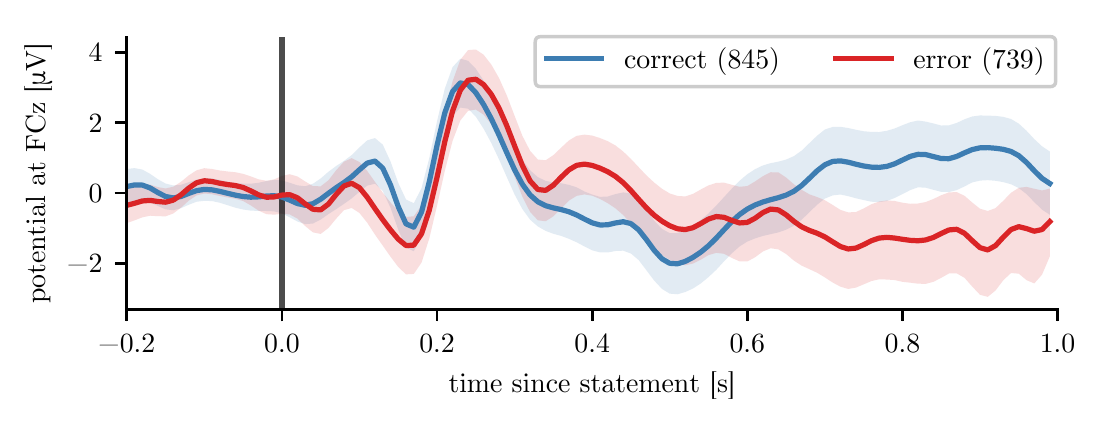}
\vspace{-3mm}
  \caption{Grand average response to preference statements for all
    \TVResultNumberKeptSubjects{} participants at frontal electrode
    FCz. Shaded areas correspond to bootstrapped 95\% confidence
    intervals of the mean.}
  \label{fig:grand-average}
\vspace{-2mm}
\end{figure}

As an introspection into the signals used for classification,
Fig.~\ref{fig:grand-average} shows the classwise average potential in
a frontocentral channel of all participants in response to the
preference statements (which is better suited for visualization due to
the stimulus alignment).  The visually evoked responses after the
statement's appearance are similar for both classes until
approximately \SI{300}{\milli\second}. However, we observed a second
positive deflection only for erroneous statements approximately
\SI{400}{\milli\second} after the statement. Such a response is
plausible since the early response will mainly depend on the visual
appearance of the stimulus and not upon the content of the text.

The discriminative response could be observed relatively early
considering the need for language processing. However, due to the
repetitive nature of the comparisons, participants likely did not have
to parse the language of the statement, but rather look for which
trajectory number was mentioned first.

The signal differences in the later part of the time window
(approximately \SIrange{600}{1000}{\milli\second}) might not solely be
explained by the statement stimulus, but could also be influenced by
the preparation of motor activity for the button press. However, the
earliest allowed occurrence of button presses was \SI{2}{\second}
after the statement (mean
\SI[separate-uncertainty,multi-part-units=single]{2.61 \pm
  0.29}{\second} across all participants), and results from a
different experimental paradigm indicate that the decoding performance
in judgment tasks does not rely solely on button press activity and
that it can be improved when reducing motor-related signal
components~\cite{kolkhorst19influence}.

\subsubsection{EEG-based Classification of Pairwise Preferences}
\label{sec:eeg-based-class}

Inspecting the classification results for individual comparisons (as
depicted in the left half of Fig.~\ref{fig:pred-results}), we achieved
a mean accuracy of \TVResultCompAccEEGStatement{} when using the
windows time-aligned to the preference statements. In the more
difficult asynchronous \emph{observation} setting---where we had no
temporal alignment to specific stimuli---we still achieved an accuracy
of \TVResultCompAccEEGObservation{}. This is especially interesting
since such information could potentially be recorded during regular
interaction with the robot, with increasing confidence after repeated
observations.

To check our expectations about the suitability of windows centered in
the trajectory (c.f., Section~\ref{sec:eeg-segmentation}), we also
evaluated two additional window alignments during observation:
Extracting time windows relative to the start or end of trajectories
rather than the center performs worse, with accuracies of
\TVResultCompAccEEGObservationStart{} and
\TVResultCompAccEEGObservationEnd{}, respectively.

Combining classifier output from both window types (accuracy of
\TVResultCompAccEEGCombined{}) yielded improvements over only using
the statement response. Note that in our evaluation setting (using
identical target trajectories across all participants),
``nonconforming'' perception by the participant (indicated by a button
accuracy of less than 1) is possible. This likely also affected the
decodable brain states and therefore the corresponding accuracies for
these trajectories.

Answering our first two questions, (1) brain signals both during
trajectory observation and after preference statements were
informative about user preferences and (2) this translated into an
accuracy of \TVResultCompAccEEGCombined{} on a single comparison level
using combined observation and statement responses.

\subsection{Results for Trajectory Ranking}
\label{sec:trajectory-ranking}

Evaluating the identification of target trajectories from pairwise
preferences, we examined 176 preference learning tasks, each
consisting of nine pairwise comparisons. The pairwise preferences were
estimated from brain signals or---for comparison with behavioral
feedback---from button presses. Our experiment design allowed us to
evaluate the ranking performance using geometric distances since we
had the ground-truth target trajectories available. For this, we used
$d_\mathrm{target}(\xi)$ as defined in Equation~\ref{eq:traj-dist} in
order to calculate the difference $\Delta d_\mathrm{target}$ between
the obtained and the best possible distance to the preference. In
addition to this absolute score, we also report the normalized
discounted cumulative gain (nDCG), which has a range from
\numrange{0}{1} and shows the relative performance based on the best
possible obtainable ranking. As the \emph{relevance} for nDCG
calculation, we used the negative distance from the preference,
shifted by the maximum. We denote the nDCG of the top-ranked item with
nDCG@1 and use nDCG@3 for the metric calculated based on the three
highest-ranked trajectories. As a baseline method, we compared our
ranking approaches to the trajectory preference perceptron proposed by
Jain~et~al.~\cite{jain15learning}, using identical features for all
methods (c.f., Section~\ref{sec:traj-rank-based}).

\subsubsection{Mean ranking performance}
\label{sec:rank-perf}

Inspecting the results of instance-based ranking with incorporated
confidences (i.e., using $\hat{R}_\mathrm{instance (conf)}$) for
different feedback types (see Fig.~\ref{fig:pred-results} and
Table~\ref{tab:results}), we observed that it is indeed possible to
successfully learn trajectory preferences from EEG-based user
feedback. For this, the use of the statement or combined EEG features
is needed. The latter achieved a mean $\Delta d_\mathrm{target}$ of
\TVResultRankDistDiffEEGCombinedInst{} for the top-ranked trajectory.

Considering the three highest-ranked trajectories, the use of
statement windows yielded an nDCG@3 of
\TVResultRankNdcgThreeEEGStatementInst{}, compared to
\TVResultRankNdcgThreeEEGCombinedInst{} for the combined features.
Limited by the lower performance on the comparison level, the ranking
based on the observation setting performed substantially worse
($\Delta d_\mathrm{target}$ of
\TVResultRankDistDiffEEGObservationInst{} and nDCG@3 of
\TVResultRankNdcgThreeEEGObservationInst{}).

Interestingly, the ranking based on combined EEG signals achieved a
similar performance as the ranking obtained by explicit button presses
(nDCG@1 of \TVResultRankNdcgOneEEGCombinedInst{} for combined EEG
vs. \TVResultRankNdcgOneButtonInst{} for button and nDCG@3 of
\TVResultRankNdcgThreeEEGCombinedInst{}
vs. \TVResultRankNdcgThreeButtonInst{}, respectively).  Note that the
improvements (which might appear counterintuitive at first) were due
to taking into account confidence values from the EEG classifiers,
which were not available for the button press (c.f., instance-based
results without confidences in Table~\ref{tab:results}). One possible
explanation for this is that for comparisons without a clear
preference, this ambivalence results in nondiscriminative brain
signals whereas the button press forces an arbitrary choice.

\begin{table*}[t]
  \centering
   \caption{Ranking performances of our methods and the baseline from
    \cite{jain15learning} on the different feedback types: EEG
    response to observations (obs), statements (stmt), and both
    combined (comb) as well as button presses (bttn). As performance
    measures we use the difference to the best obtainable
    $d_\mathrm{target}$ (lower is better) and the normalized
    discounted cumulative gain (nDCG, higher is better) for the
    top-ranked (nDCG@1) and the three highest-ranked trajectories
    (nDCG@3).}
  \label{tab:results}
  \vspace{-2mm}
  \begin{tabular}{lllllllllllll}
\toprule
{} & \multicolumn{4}{c}{$\Delta d_{\mathrm{target}}$} & \multicolumn{4}{c}{nDCG@1} & \multicolumn{4}{c}{nDCG@3} \\
\textbf{feedback type} &                                obs &                                stmt &                                comb &                                bttn &                                obs &                               stmt &                               comb &                               bttn &                                obs &                               stmt &                               comb &                               bttn \\
\textbf{ranking                               } &                                    &                                     &                                     &                                     &                                    &                                    &                                    &                                    &                                    &                                    &                                    &                                    \\
\midrule
\textbf{traj. perceptron \cite{jain15learning}} &          \num{0.15442821478579546} &           \num{0.11232030711647728} &           \num{0.11118673220113638} &           \num{0.07608196314488636} &           \num{0.6369954312184859} &           \num{0.7232053685274237} &           \num{0.7435419892142382} &           \num{0.8117342184312671} &           \num{0.6787036895159893} &           \num{0.7185014258920347} &           \num{0.7322863417434063} &           \num{0.7715267218590668} \\
\textbf{instance                              } &           \num{0.1323501544130682} &           \num{0.10522433629488637} &           \num{0.10456309424431819} &           \num{0.07786819587499999} &           \num{0.6796330965652877} &           \num{0.7428283343170193} &           \num{0.7523756183278117} &           \num{0.8120623738951509} &           \num{0.6771768719694443} &           \num{0.7271577995237026} &           \num{0.7466252944021187} &           \num{0.8151521986687196} \\
\textbf{instance (conf)                       } &           \num{0.1323501544130682} &  \textbf{\num{0.07410801775056818}} &  \textbf{\num{0.07316602942897726}} &           \num{0.07786819587499999} &           \num{0.6796330965652877} &  \textbf{\num{0.8196026968061509}} &  \textbf{\num{0.8223883830620982}} &           \num{0.8120623738951509} &           \num{0.6771768719694443} &  \textbf{\num{0.8209289712513239}} &  \textbf{\num{0.8406990733580636}} &           \num{0.8151521986687196} \\
\textbf{feature                               } &  \textbf{\num{0.1056308666846591}} &           \num{0.08943836872840909} &           \num{0.09161111971420453} &  \textbf{\num{0.05277182304261365}} &  \textbf{\num{0.7395249586478958}} &           \num{0.7774566159664412} &           \num{0.7871701131093086} &  \textbf{\num{0.8686028208150639}} &  \textbf{\num{0.7348020215084017}} &           \num{0.7756991825198994} &           \num{0.7808941673987014} &  \textbf{\num{0.8467921461892186}} \\
\bottomrule
  \end{tabular}
\vspace{-3mm}
\end{table*}

\newpage

Using our feature-based trajectory ranking approach
($\hat{R}_\mathrm{feat}$, where training labels are based on user
feedback), we could observe that for low label noise (i.e., button
presses), it outperformed all other ranking settings (mean
$d_\mathrm{target}$ of \TVResultRankDistDiffButtonFeat{} and a mean
nDCG@1 of \TVResultRankNdcgOneButtonFeat{}). Since label uncertainty
in the form of comparison confidences is, however, not incorporated in
this approach, it performed worse than the instance-based ranking when
utilizing the less accurate EEG-based feedback. In the adequate
comparison with the instance-based approach without confidences, the
feature-based ranking performed better. As shown in
Table~\ref{tab:results}, our feature-based approach outperformed the
trajectory preference perceptron proposed in \cite{jain15learning} in
all feedback settings.

\subsubsection{Analysis of Results for Individual Participants and
  Preferences}
\label{sec:analys-results-indiv}

\begin{figure}[t]
  \centering
  \includegraphics[width=\columnwidth]{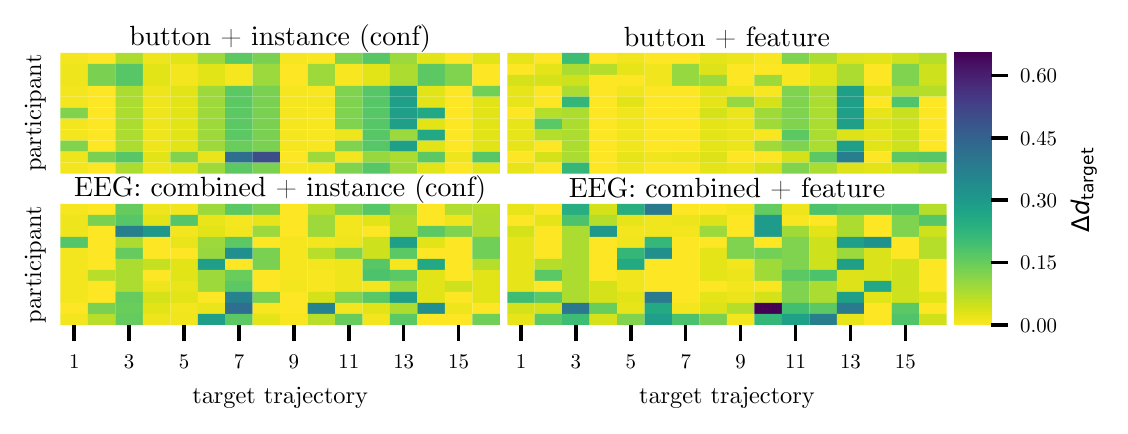}

  \caption{Ranking performance for individual participants and
    preferences: The top matrices are based on button press and the
    bottom matrices are based on the combined EEG responses. The left
    matrices use instance-based ranking without additional trajectory
    information and the right ones use the feature-based ranking using
    trajectory and scene features. Entries in each matrix correspond
    to a single participant (row) for a single target trajectory
    (column). Performances are measured using the difference from the
    target $\Delta d_\textrm{target}$ for the top-ranked trajectory,
    lower is better)}
  \label{fig:blockwise-performance}
\end{figure}

Performing identical experiments with all participants allowed us to
gain insight into the influence of different tasks and users on the
ranking performance. The heat maps in
Fig.~\ref{fig:blockwise-performance} indicate the performance of
participants for the different preference tasks. Looking at the button
results with low feedback noise in the top two matrices, the
similarities within each individual column show that performance
variations were systematic and depended on the set of trajectories and
targets.

Specifically, we observed comparatively low performance for target
trajectories 3 and 13. Here, also the button feedback accuracy (not
shown) is lower for most participants, indicating ``harder''
comparisons that also translated into worse EEG-based decoding and
ranking performance. While behavioral performance is similar across
most participants, we observed a higher variance in the EEG-based
setting due to the inherent interperson variability in EEG
measurements. However, a recovery of preferences was possible in most
of the preference tasks for all participants. %

Answering question (3), ranking based on EEG-based predictions was
possible both in the instance-based and the feature-based
setting. Moreover, we found that utilizing EEG-based confidence values
in addition to the pairwise preference prediction allowed a ranking
performance based on brain signals that is comparable to results
obtained by button presses.

\section{Conclusion}
\label{sec:conclusion}

In this work, we presented a novel approach to learn user preferences
for trajectories from brain signals based on pairwise comparisons. Our
approach predicts pairwise preferences from EEG data both during
passive observation of trajectories and in response to explicit
statements. We utilize these predictions
to rank observed trajectories.

In extensive experiments, we demonstrated that brain signals as a
feedback modality were informative about a user's target trajectory
and the latter could be reliably predicted in a pairwise
setting. Furthermore, we showed that---despite the low signal-to-noise
ratio of EEG signals---ranking trajectories using the EEG-based
pairwise preference predictions allowed us to identify the target
trajectories with a performance comparable to explicit button presses.

Our results open up paths for future work both on EEG-based active
learning of
preferences~\cite{sadigh17active,akrour14programming}---where feedback
could also be repeated on demand to reduce measurement noise---and on
utilizing brain signals in a passive way during human--robot
interaction to improve robotic behavior without explicitly querying
the human.

\addtolength{\textheight}{-6.5cm}
\section*{Acknowledgment}
\begin{small}
  
  The authors would like to thank Robin Burchard for help recording
  the robot trajectories and Joseline Veit for help conducting the EEG
  experiments. This work was (partly) supported by
  BrainLinks-BrainTools, Cluster of Excellence funded by the German
  Research Foundation (DFG, grant number EXC 1086). Additional support
  was received from the German Federal Ministry of Education and
  Research under grant OML, the DFG through INST 39/963-1 FUGG as well
  as the Ministry of Science, Research and the Arts of
  Baden-W\"urttemberg for bwHPC.

\end{small}
\bibliographystyle{IEEEtran}
\bibliography{literature}
\end{document}